%% file: 0_main.tex
\documentclass[10pt,twocolumn,letterpaper]{article}

\usepackage{makecell}
\usepackage{rotating}
\usepackage{tabularray}
\usepackage{ragged2e}
\usepackage{lipsum}
\usepackage{bbm}
\usepackage{enumitem}
\usepackage{verbatim}
\usepackage{bm}
\usepackage{flushend}
\usepackage{setspace}
\usepackage{cuted}
\usepackage{hhline}
\usepackage{tabularx}
\usepackage{authblk}
\usepackage[font=small,labelfont=bf]{caption}

\usepackage{iccv}
\usepackage{times}
\usepackage{epsfig}
\usepackage{graphicx}
\usepackage{amsmath}
\usepackage{amssymb}

\usepackage{multicol}
\usepackage{microtype}
\usepackage{booktabs}       %
\usepackage{multirow}
\usepackage{wrapfig}

\usepackage{bm}
\usepackage{adjustbox}
\usepackage{array}
\usepackage{pifont}
\usepackage{float}
\usepackage{xfrac}
\usepackage{colortbl}
\usepackage[normalem]{ulem}
\usepackage{mathtools}
\usepackage{breqn}
\usepackage{stmaryrd}

\usepackage{caption}
\usepackage{subcaption}

\usepackage{comment}
\usepackage[dvipsnames,table]{xcolor}
\usepackage{soul}
\usepackage{cite}

\input{abbrev.tex}

\usepackage[pagebackref=true,breaklinks=true,letterpaper=true,colorlinks,citecolor=blue,bookmarks=false]{hyperref}

\iccvfinalcopy %

\def\iccvPaperID{x} %

\usepackage{xcolor}

\newcommand{\Xuanlong}{\textcolor{black}}
\newcommand\gianni{\textcolor{black}}
\begin{document}

\title{The Robust Semantic Segmentation UNCV2023 Challenge Results}

\makeatletter
\renewcommand\AB@affilsepx{, \protect\Affilfont}
\makeatother

\author[1,2]{Xuanlong Yu}

\author[3]{Yi Zuo}
\author[3]{Zitao Wang}
\author[3]{Xiaowen Zhang}
\author[3]{Jiaxuan Zhao}
\author[3]{Yuting Yang}
\author[3]{Licheng Jiao}

\author[3]{Rui Peng}
\author[3]{Xinyi Wang}
\author[3]{Junpei Zhang}
\author[3]{Kexin Zhang}
\author[3]{Fang Liu}

\author[4]{Roberto Alcover-Couso}
\author[4]{Juan C. SanMiguel}
\author[4]{Marcos Escudero-Viñolo}

\author[5]{Hanlin Tian}
\author[5]{Kenta Matsui}
\author[6]{Tianhao Wang}
\author[5]{Fahmy Adan}

\author[7]{Zhitong Gao}
\author[7]{Xuming He}

\author[8]{Quentin Bouniot}
\author[8]{Hossein Moghaddam}

\author[9]{Shyam Nandan Rai}
\author[9]{Fabio Cermelli}
\author[9]{Carlo Masone}

\author[10]{Andrea Pilzer}
\author[11]{Elisa Ricci}
\author[12]{Andrei Bursuc}
\author[13]{Arno Solin}
\author[13]{Martin Trapp}
\author[13]{Rui Li}
\author[14]{Angela Yao}
\author[5]{Wenlong Chen}
\author[15]{Ivor Simpson}
\author[16]{Neill D. F. Campbell}
\author[2]{Gianni Franchi}

\affil[1]{SATIE, Paris-Saclay University, France}
\affil[2]{U2IS, ENSTA Paris, Institut Polytechnique de Paris, France}

\affil[3]{Key Laboratory of Intelligent Perception and Image Understanding of the Ministry of Education, Xidian University, China}

\affil[4]{VPU-Lab, Autonomous University of Madrid (UAM), Spain}
\affil[5]{Imperial College London, UK}
\affil[6]{The University of Texas at Dallas, USA}

\affil[7]{ShanghaiTech University, China}

\affil[8]{LTCI, Télécom Paris, Institut Polytechnique de Paris, France}

\affil[9]{Politecnico di Torino, Italy}

\affil[10]{NVIDIA AI Technology Center, Italy}
\affil[11]{University of Trento, Italy}
\affil[12]{valeo.ai, France}
\affil[13]{Aalto University, Finland}
\affil[14]{National University of Singapore, Singapore}
\affil[15]{University of Sussex, UK}
\affil[16]{University of Bath, UK}


\maketitle

\begin{abstract}
This paper outlines the winning solutions employed in addressing the MUAD uncertainty quantification challenge held at ICCV 2023. The challenge was centered around semantic segmentation in urban environments, with a particular focus on natural adversarial scenarios.
The report presents the results of 19 submitted entries, with numerous techniques drawing inspiration from cutting-edge uncertainty quantification methodologies presented at prominent conferences in the fields of computer vision and machine learning and journals over the past few years. Within this document, the challenge is introduced, shedding light on its purpose and objectives, which primarily revolved around enhancing the robustness of semantic segmentation in urban scenes under varying natural adversarial conditions. The report then delves into the top-performing solutions.
Moreover, the document aims to provide a comprehensive overview of the diverse solutions deployed by all participants. By doing so, it seeks to offer readers a deeper insight into the array of strategies that can be leveraged to effectively handle the inherent uncertainties associated with autonomous driving and semantic segmentation, especially within urban environments.
\end{abstract}

\section{Introduction}\label{sec:intro}
As computer vision applications based on Deep Neural Networks (DNN) penetrate various aspects of our lives, ensuring reliable and robust performance becomes paramount. 
This is particularly evident in domains like autonomous driving, where DNNs play a pivotal role in comprehending the surrounding environment. However, the potential for inaccuracies due to various sources of uncertainty looms large. The quantification of uncertainty inherent in DNN predictions holds significant importance. Inaccurate assessments of uncertainty could lead to erroneous decisions, a scenario that could be catastrophic, as witnessed in cases like autonomous driving~\cite{li2016trolley,kone2020safety}. \gianni{An increasing number of datasets are emerging that could model a broader spectrum of potential uncertain scenarios~\cite{sakaridis2021acdc,shift2022}.}
Nonetheless, 
the datasets served for the DNNs may not adequately encapsulate the vast array of conditions and objects prevalent in real-world scenarios. Autonomous driving, for instance, contends with adversities weather conditions like inclement snow, rain, and fog, and may contend the presence of Out-of-Distribution (OOD) instances.
By delving into the study and quantification of uncertainty, we attain invaluable insights into areas where the model may be less confident, guiding us to adopt precautionary measures or invoke human intervention as necessary.

Thanks to the release of relative datasets~\cite{pinggera2016lost,yu2020bdd100k,chan2021segmentmeifyoucan,Franchi2022MUAD}, significant progress has been achieved in recent years on the OOD example detection in semantic segmentation~\cite{di2021pixel, liang2018enhancing, besnier2021triggering,tian2022pixel, grcic2021dense, lis2019detecting,kendall2015bayesian,franchi2022latent}.
To advance the field of semantic segmentation robustness and uncertainty quantification, the first ACDC Challenge~\cite{sakaridis2021acdc}, which was held at Vision for All Seasons workshop in IEEE / CVF Computer Vision and Pattern Recognition Conference (CVPR 2022), aims to deal with semantic segmentation under complex weather conditions and changes in the visual description of objects caused by weather. Yet OOD detection task is not covered in the challenge. 
Meanwhile, compared with the previous anomaly segmentation benchmarks such as SegmentMeIfYouCan~\cite{chan2021segmentmeifyoucan}, MUAD~\cite{Franchi2022MUAD}, a larger-scale dataset for autonomous driving with multiple uncertainty types, is proposed to evaluate the robustness of the algorithms in both segmentation accuracy and uncertainty quantification quality, especially OOD detection.

Inspired by the previous works, in this paper, we will introduce the MUAD challenge~\footnote{\url{https://codalab.lisn.upsaclay.fr/competitions/8007}}, which is held at the ICCV 2023 UNCV workshop. The challenge is launched on the CodaLab~\cite{codalab_competitions_JMLR} platform. The researchers \Xuanlong{were}
encouraged to submit both the semantic segmentation class prediction maps and the corresponding prediction confidence maps of a subset selected from MUAD test sets. According to the final ranking with respect to the semantic segmentation and OOD detection performance, we select the best solutions and will go into the details in the later sections. We believe that this challenge is meaningful for the progress of safety AI.

\section{The MUAD challenge}\label{sec:challenge}
\subsection{The dataset}
MUAD~\cite{Franchi2022MUAD}, a synthetic dataset pioneering autonomous driving research, introduces various uncertainties in the real world. It offers 10413 instances encompassing day and night scenes: 3420 in the training set, 492 in the validation set, and 6501 in the test set. There are seven subsets in the test set, including normal set, OOD set, low adversity and high adversity sets, etc. Only test sets contain the OOD objects or special weather conditions such as rain, fog, or snow with different severities. We randomly select the images from the test to conduct the challenge test set. 

The participants download the training and validation sets (containing the RGB images and the corresponding ground truth maps) as well as the test set (only the RGB images are provided), then design and train the models. Class prediction maps and confidence maps are required to be submitted during evaluation, which aims to provide enough information to help the decision-maker identify the patterns and find out the uncertain prediction like OOD objects. Different levels of weather conditions will also be challenging to the robustness of the models and the OOD example detectors.

\subsection{The evaluation protocols}

We evaluate the overall semantic segmentation performance of the algorithms using the mean Intersection over Union (mIoU). For the confidence maps, we choose the mean Expected Calibration Error
(mECE)~\cite{naeini2015obtaining} for the calibration of uncertainties and measure the OOD detection performance using mean Areas Under the operating Curve (mAUROC) and mean Areas Under the Precision/Recall curve (mAUPR), as well as mean False Positive Rate at 95$\%$ recall (mFPR), similarly to~\cite{hendrycks17baseline}. All the mean is calculated across all pixels. For mIoU, mAUROC and mAUPR, the higher is better. For mECE and mFPR, the lower is better. 

\input{1_challenge_results}

\section{Conclusion}
\Xuanlong{Results of UNCV2023 challenge were presented. The main theme of the challenge is to evaluate the uncertainty quantification performance of the semantic segmentation models based on the MUAD dataset. The goal is to discover uncertainty quantification solutions in autonomous driving scenarios and assess the impact of different sources of uncertainty on model performance.}

\gianni{As per the solutions submitted by participants, a variety of model architectures, backbones, data augmentation techniques, and model ensembles emerged as prevalent choices among most teams. Additionally, approaches involving region normalization, novel loss functions, test time batch norm adaptation, and model calibration were also brought into the spotlight. These diverse strategies collectively offer valuable insights into the practical deployment of robust semantic segmentation algorithms and the concurrent quantification of uncertainty within intricate urban environments.}

\gianni{We view these approaches as a repository of strategies akin to a ``bag of tricks". The prospect of optimizing the synergy among these methods and exploring potential conflicts between them holds significant promise for future research endeavors.}

\section*{Acknowledgements}
We gratefully acknowledge the support of DATAIA Paris-Saclay institute and AID Project ACoCaTherm which supported the creation of the dataset.

\clearpage

\bibliographystyle{ieee_fullname}
\bibliography{references}

\end{document}

%% file: abbrev.tex
\newcommand{\vtheta}{\boldsymbol{\theta}}

\newcommand{\vy}{\mathbf{y}}
\newcommand{\vx}{\mathbf{x}}

\definecolor{dgreen}{rgb}{0.0, 0.5, 0.0}
\definecolor{better}{rgb}{0.19, 0.55, 0.91}
\definecolor{worse}{rgb}{0.82, 0.1, 0.26}

\newcommand{\second}{\cellcolor[rgb]{0.847,0.882,0.949}}
\newcommand{\first}{{\cellcolor[rgb]{0.557,0.663,0.859}}}

\DeclareMathOperator*{\argmax}{arg\,max}

%% file: 1_challenge_results.tex
\section{The MUAD challenge results}\label{sec:results}

\subsection{Notations}

For the training of segmentation algorithms, all the process commences with a dataset of training examples denoted as $\mathcal{D}=\{\vx_i, \vy_i\}_{i=1}^{|\mathcal{D}|}$, where $|\mathcal{D}|$ is the total number of image-label pairs. These pairs consist of images $\vx_i\in\mathbb{R}^{C\times H\times W}$ and corresponding label maps $\vy_i\in \llbracket 0, N_C \llbracket^{H\times W}$, treated as realizations of a joint distribution $\mathcal{P}{(X,Y)}$. In this context, $C$, $H$, and $W$ denote the \gianni{number of} channel, %
height, and width of the image, respectively, while $N_C$ represents the %
\gianni{number of} classes within the dataset.

The neural network $F_{\vtheta}(\cdot)$ is then employed to process the input data $\vx_i$. This network functions as a parametric probabilistic model, denoted by $\hat{\vy}_i = F_{\vtheta}(\vx_i) = P(Y=\vy_i | X=\vx_i ; \vtheta)$. In essence, it estimates the conditional probability of the label $\vy_i$ given the input image $\vx_i$ based on a set of parameters $\vtheta$. %

\subsection{Overview of the results}
\noindent\textbf{Baseline\quad}
\gianni{The baseline model is established using DeepLabV3+\cite{chen2018encoder}, which employs a ResNet101\cite{he2016deep} as its underlying architecture. The training protocol adheres to the official MUAD configuration~\cite{Franchi2022MUAD}.
Let $\hat{\mathbf{y}}^{\text{[pred]}}_i$ symbolize the class prediction map generated by selecting the class with the highest value along the classes of $\hat{\mathbf{y}}_i$. This can be formally expressed as:
$\hat{y}^{\text{[pred]}}_i(h,w) = \argmax_c \hat{y}_i(c,h,w)$.
Subsequently, the confidence map $\hat{\mathbf{y}}^{\text{[conf]}}_i$ is based on the associated maximum probability values\Xuanlong{~\cite{hendrycks17baseline}}:
$\hat{y}^{\text{[conf]}}_i(h,w) = \max_c \hat{y}_i(c,h,w)$.}

\begin{figure*}[t]
\centering
\begin{tabular}{ccccc}
Image & Groundtruth & Baseline & 2nd place solution & 1st place solution\\
\includegraphics[width=0.18\linewidth]{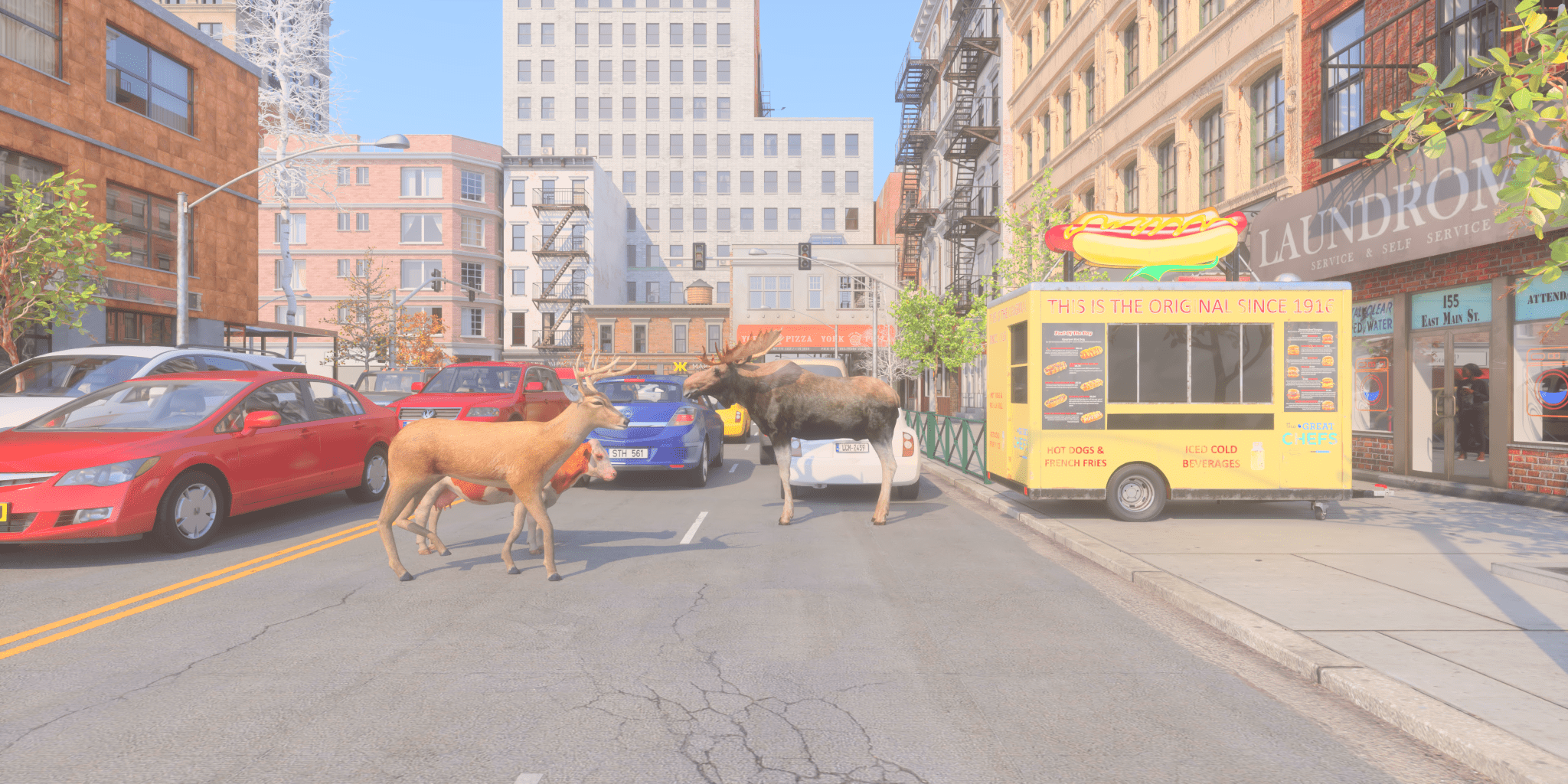}&
\includegraphics[width=0.18\linewidth]{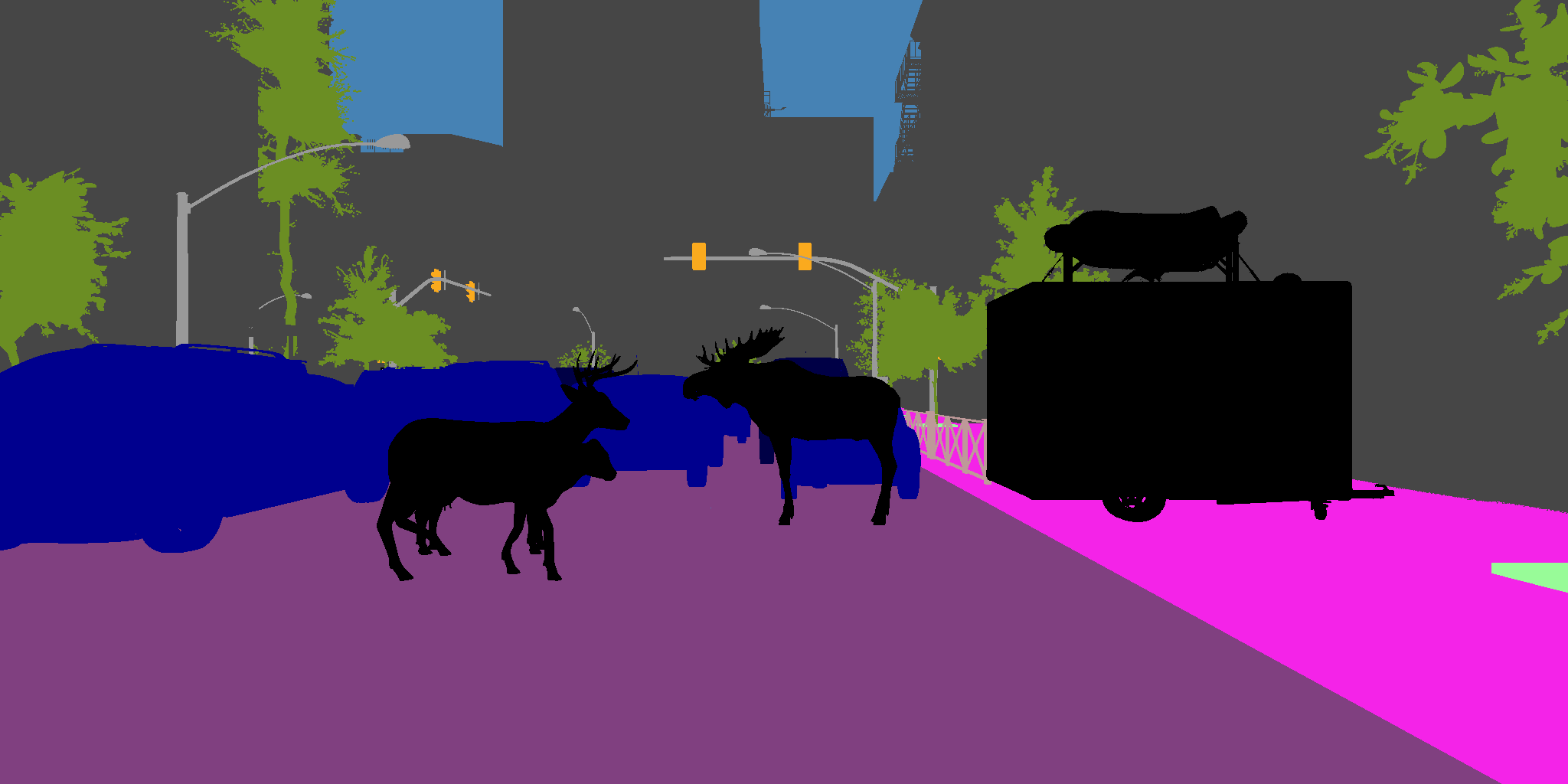}&
\includegraphics[width=0.18\linewidth]{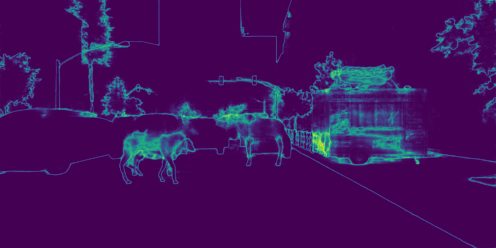}&
\includegraphics[width=0.18\linewidth]{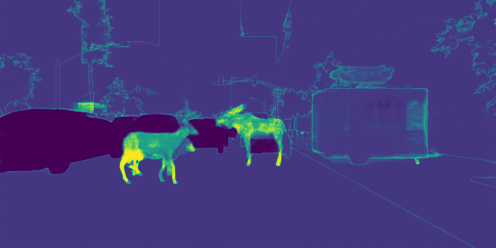}&
\includegraphics[width=0.18\linewidth]{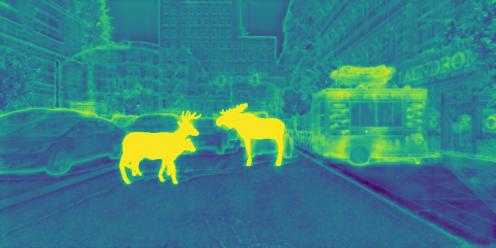}\\
\includegraphics[width=0.18\linewidth]{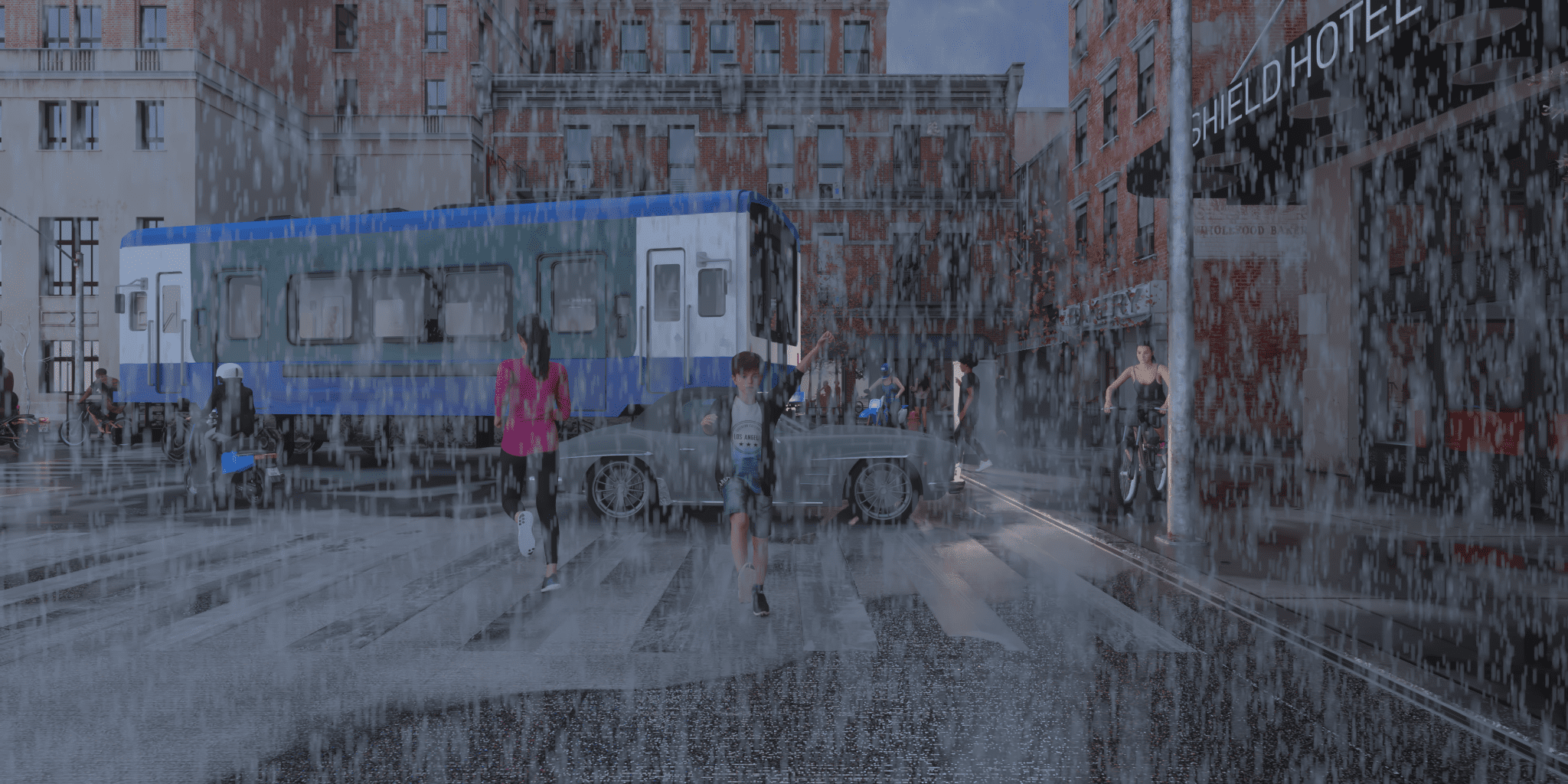}&
\includegraphics[width=0.18\linewidth]{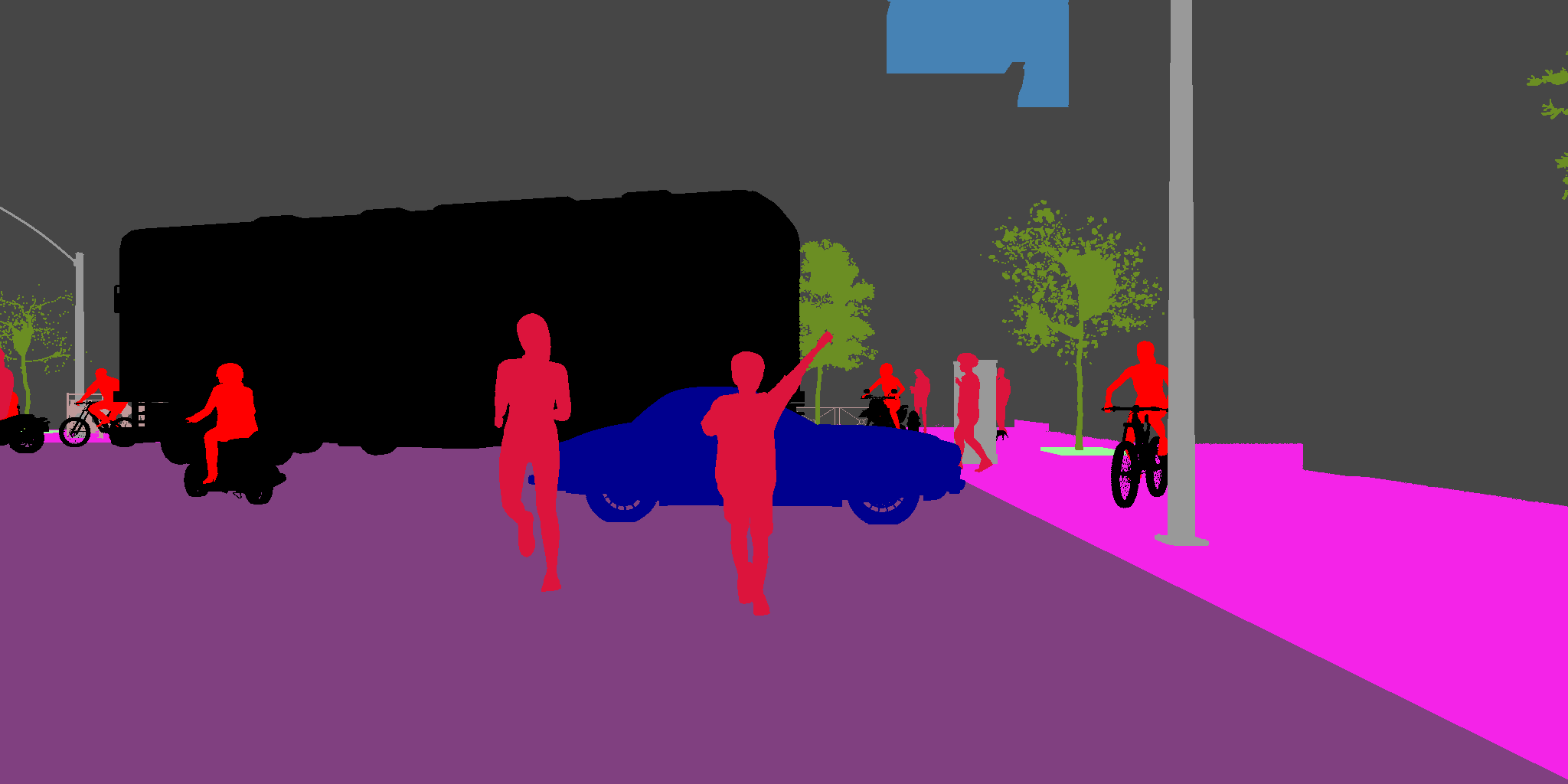}&
\includegraphics[width=0.18\linewidth]{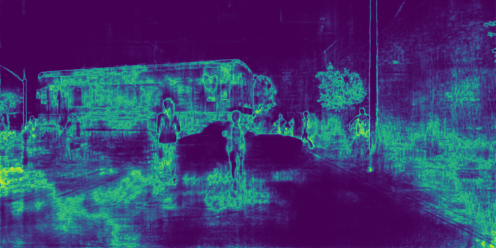}&
\includegraphics[width=0.18\linewidth]{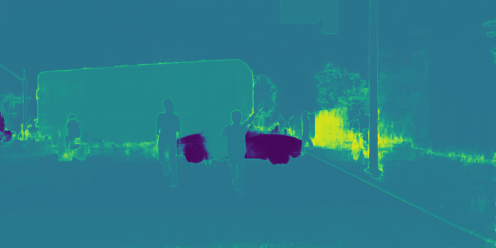}&
\includegraphics[width=0.18\linewidth]{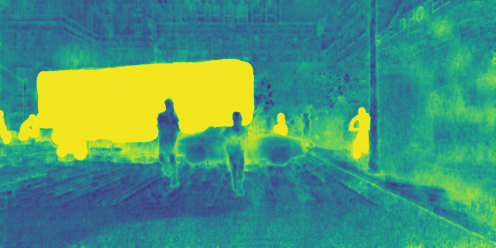}\\
\includegraphics[width=0.18\linewidth]{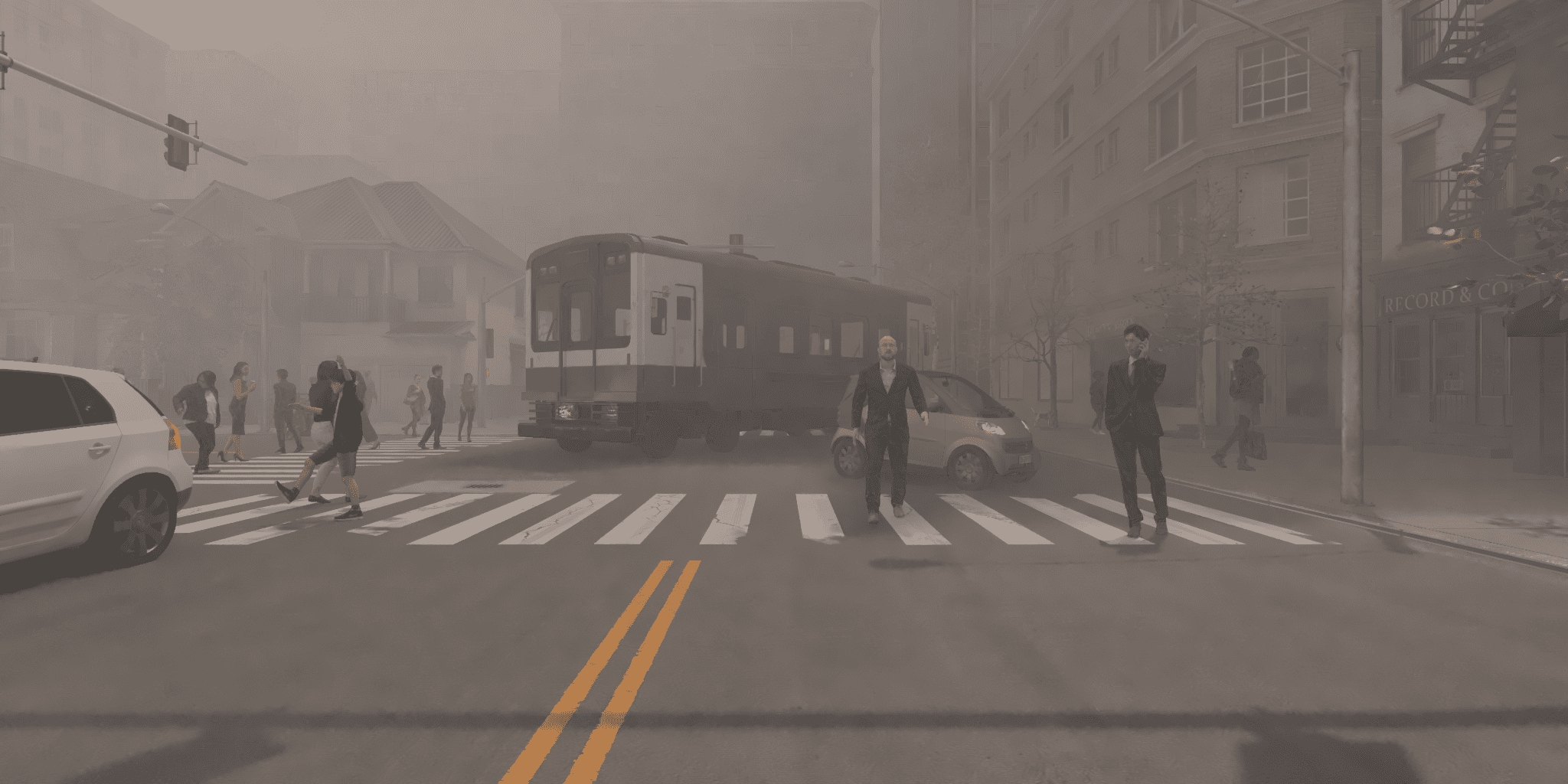}&
\includegraphics[width=0.18\linewidth]{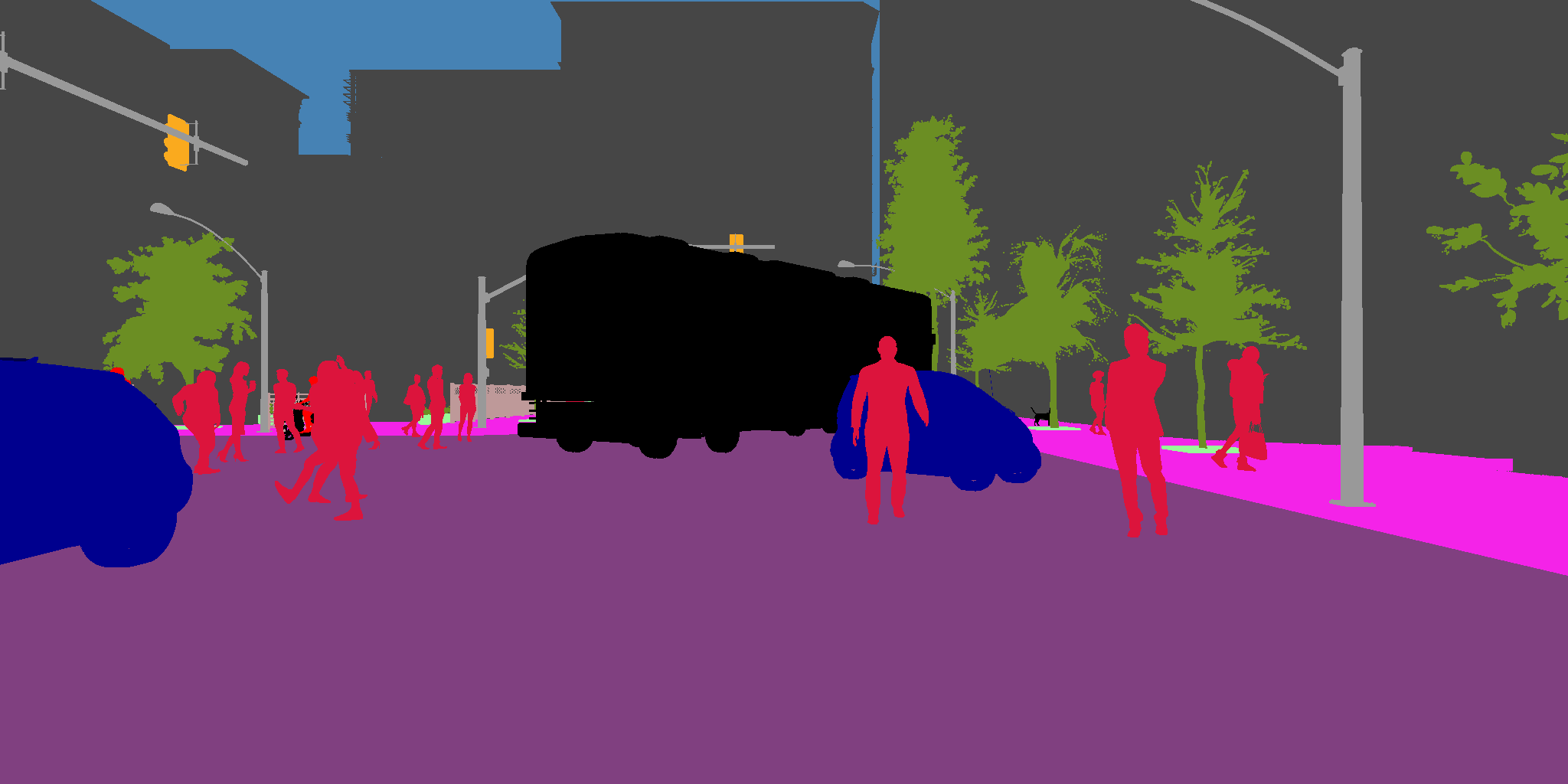}&
\includegraphics[width=0.18\linewidth]{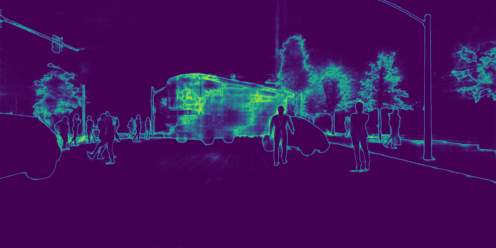}&
\includegraphics[width=0.18\linewidth]{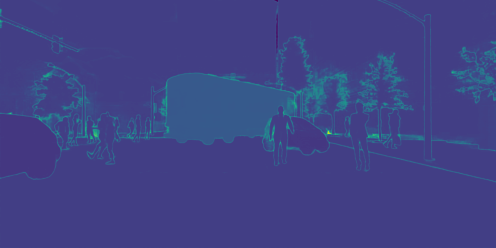}&
\includegraphics[width=0.18\linewidth]{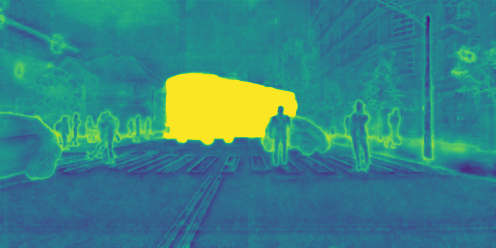}\\
\includegraphics[width=0.18\linewidth]{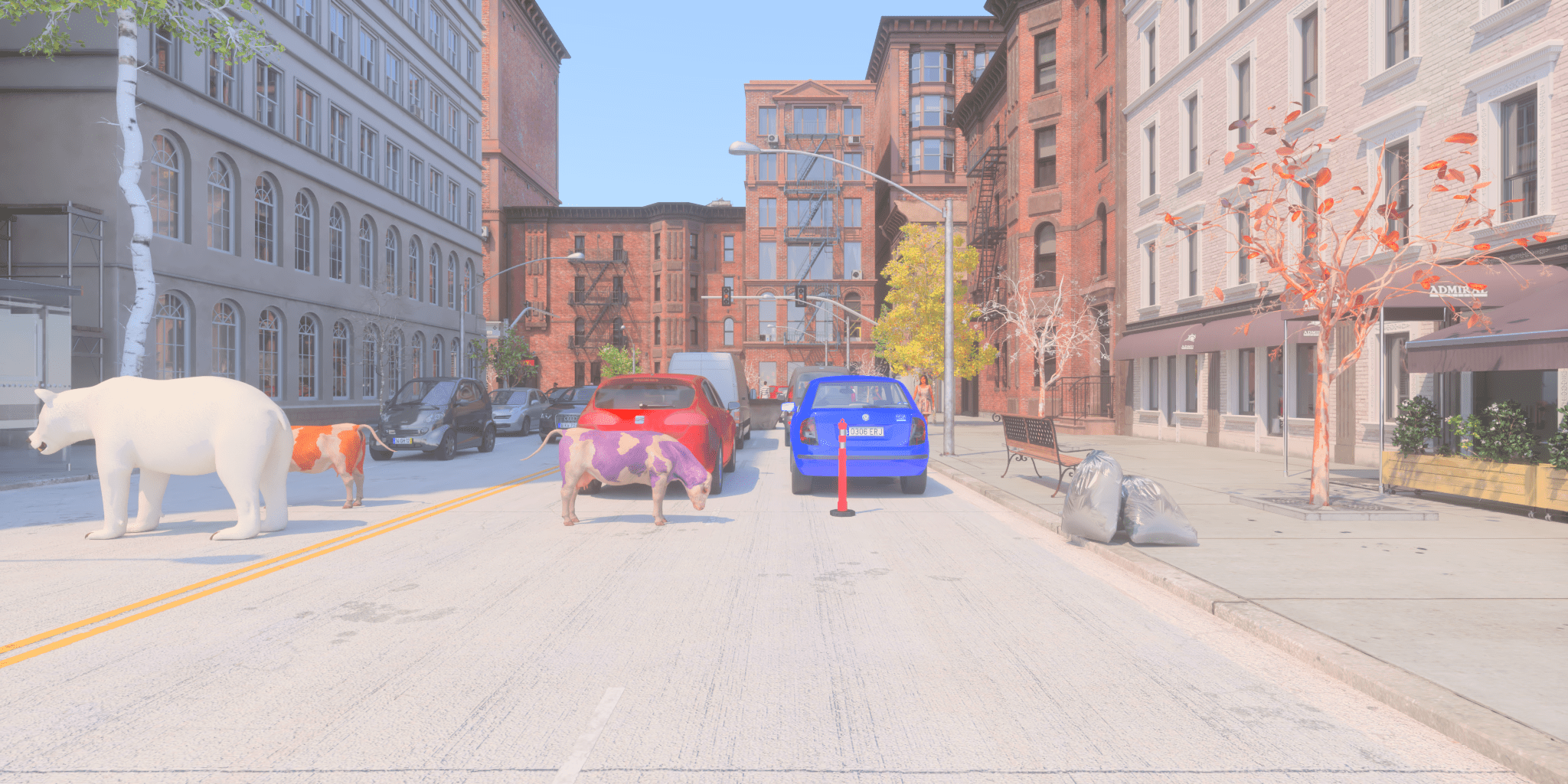}&
\includegraphics[width=0.18\linewidth]{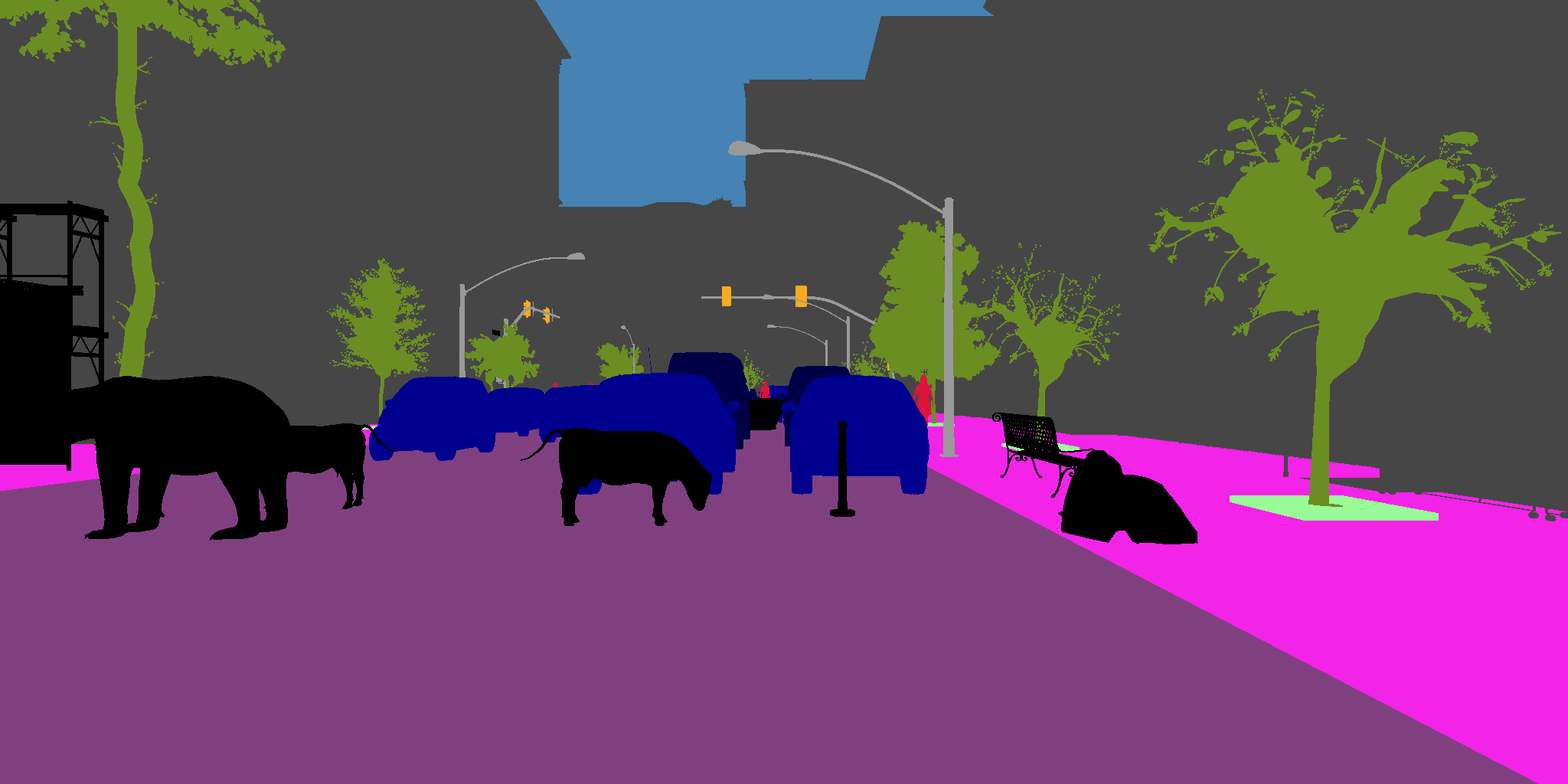}&
\includegraphics[width=0.18\linewidth]{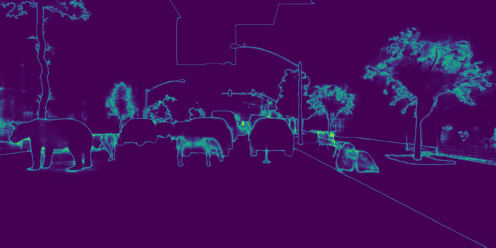}&
\includegraphics[width=0.18\linewidth]{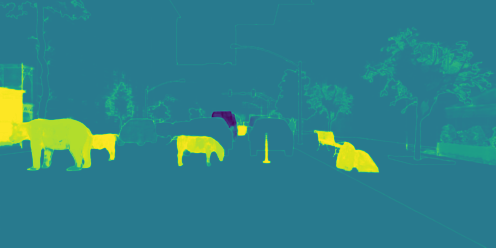}&
\includegraphics[width=0.18\linewidth]{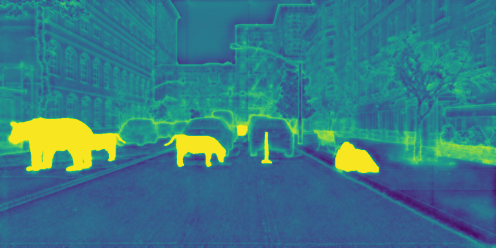}
\\
\end{tabular}
\vspace{0.3em}
\caption{\textbf{Illustration of confidence maps on one image of MUAD.} The latter three columns are the confidence maps given by the baseline, the 2nd place solution, and the 1st place solution, respectively. The brighter, the more uncertain the corresponding class prediction is.  Note that the class \texttt{train}, \texttt{bicycle/motorcycle}, \texttt{stone}, \texttt{stand food}, and the \texttt{animals} are OOD examples, which are visualized as black areas on the groundtruth maps.}
\label{fig:vis}
\end{figure*}

\noindent\textbf{Overview of the results\quad}
\gianni{By the end of the challenge,  a total of 73 teams registered to participate. Throughout the course of the challenge, more than 700 results were submitted to the CodaLab server, with 19 teams opting to disclose their rankings publicly. The conclusive outcomes are presented in Table~\ref{tab:overall_results}, where the most successful two methods are distinctly highlighted. The ``NA" designation indicates instances where mECE metrics were impacted by unnormalized confidence scores. For the ultimate ranking, we utilize the mAUROC results, recognizing that the solutions might involve trade-offs across various metrics. It is apparent that the majority of teams have outperformed the baseline, particularly the top 9 teams, which have demonstrated notable advancements across nearly all metrics.}
\Xuanlong{Figure~\ref{fig:vis} shows the visualizations on the confidence maps given by the baseline, the 2nd place solution, and the 1st place solution. When different weather conditions occur in the input images, the 1st place solution can much better highlight the OOD examples than the baseline confidence map.}

\begin{figure*}[t!h]
    \centering
\begin{tabular}{cc}
\includegraphics[width=0.5\linewidth]{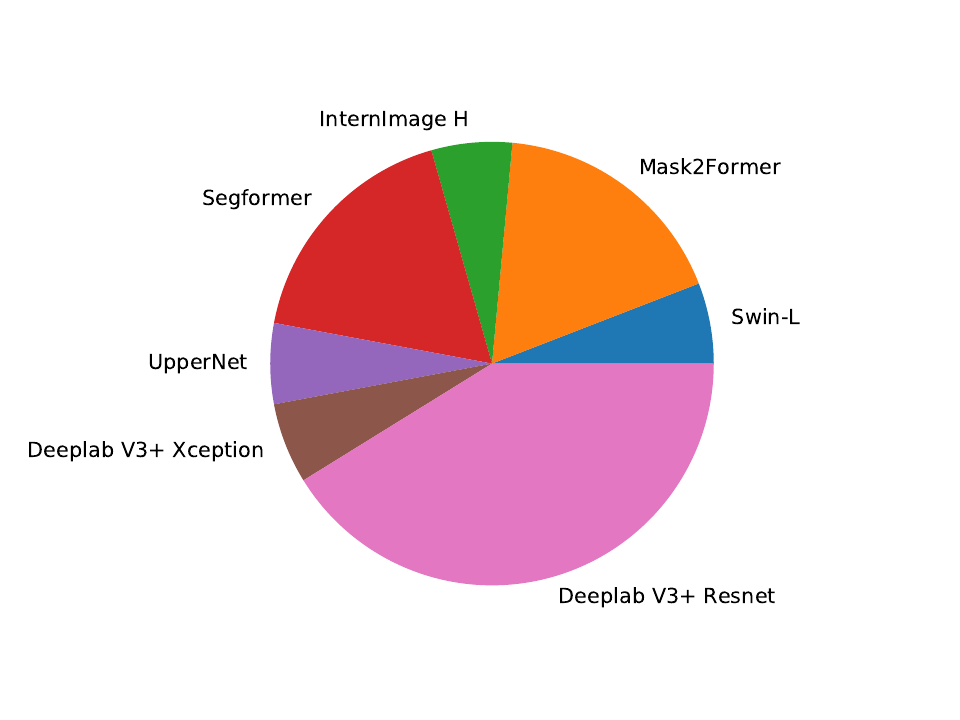}
\includegraphics[width=0.5\linewidth]{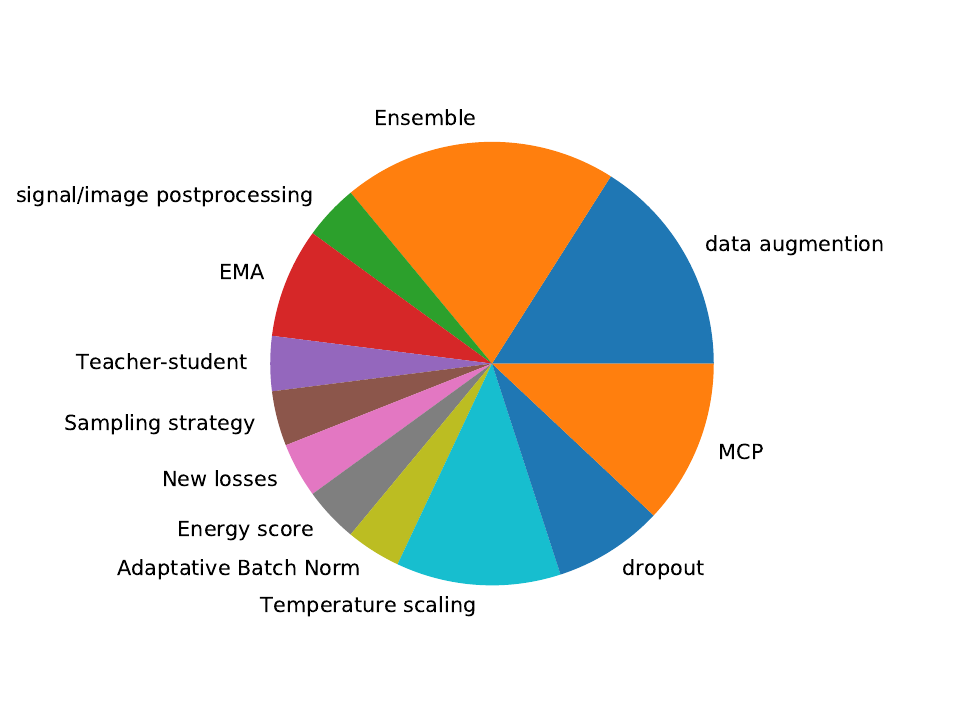}
\end{tabular}
    \caption{\textbf{The statistics of the different solutions.} The pie chart on the left displays a visualization of the diverse architectures employed throughout the challenge, while the chart on the right showcases the array of solution types embraced by the participants. }
    \label{fig:diagram}
\end{figure*}

\noindent\textbf{Overview on the submitted solutions\quad}
\gianni{In their pursuit of innovative solutions, a significant majority of participants opted for the Deeplab V3+ architecture coupled with Resnet101. However, several participants proposed to modify the architecture or utilized multiple DNNs, as visualized in Figure \ref{fig:diagram}. Furthermore, competitors employed a diverse array of strategies, such as Energy Score, Biased Class disagreement, and temperature scaling approaches. Most prominently, ensembling strategies were extensively employed, accompanied by appropriate data augmentation methodologies aimed at enhancing DNN robustness. For a comprehensive view of the participants' statistical breakdown, refer to Figure \ref{fig:diagram}. Furthermore, detailed insights into the distinct solutions adopted can be found in Section \ref{sec:topsolutions}.}

\begin{table}[t]
\centering
\scalebox{0.69}{
\begin{tabular}{rlccccc} 
\toprule
\# & User Name & mAUROC $\uparrow$ & mAUPR $\uparrow$ & mFPR $\downarrow$ & mECE $\downarrow$ & mIoU $\uparrow$ \\ 
\toprule
\toprule
1 & zuoyi & \first \textbf{0.9516} & 0.6053 & \second 0.1722 & NA & 0.4887 \\
2 & arui & \second 0.9515 & \first \textbf{0.6379} & \first \textbf{0.1499} & 0.0726 & 0.5123 \\
3 & bee & 0.9149 & \second 0.6072 & 0.2230 & 0.0698 & \second 0.5340 \\
4 & MahouShoujo & 0.9094 & 0.3520 & 0.2391 & NA & 0.3837 \\
5 & rac & 0.8510 & 0.4435 & 0.3983 & \second 0.0603 & \first \textbf{0.6454} \\
6 & zachtian & 0.8393 & 0.2196 & 0.3850 & 0.1075 & 0.3274 \\
7 & Tonnia & 0.8393 & 0.2745 & 0.4196 &  NA & 0.4599 \\
8 & drop08 & 0.8356 & 0.2502 & 0.3898 & 0.0942 & 0.4558 \\
9 & Shyam671 & 0.8341 & 0.3038 & 0.3091 & \first \textbf{0.0531} & 0.3817 \\ 
10 & xtz & 0.8280 & 0.4767 & 0.3536 & 0.1540 & 0.5049 \\
11 & Team\_Rocket & 0.7673 & 0.1821 & 0.4720 & 0.1855 & 0.3796 \\
12 & h.m & 0.7644 & 0.2031 & 0.4512 & 0.1657 & 0.4134 \\
13 & HPeter & 0.7429 & 0.1712 & 0.5204 & 0.3116 & 0.3656 \\
14 & Janes\_migadel & 0.7418 & 0.1806 & 0.5880 & 0.1799 & 0.2848 \\
15 & Yuehua.DING & 0.7411 & 0.1641 & 0.5020 & 0.2165 & 0.3668 \\
16 & dens03 & 0.7399 & 0.1884 & 0.4787 & 0.2521 & 0.3764 \\
\midrule
17 & Baseline & 0.7337 & 0.1790 & 0.5253 & 0.2880 & 0.3690 \\
\midrule
18 & NathLaMenace & 0.7337 & 0.1790 & 0.5253 & 0.2880 & 0.3690 \\
19 & Sp4n & 0.7091 & 0.1907 & 0.7499 & 0.0970 & 0.5309 \\
\bottomrule
\\
\end{tabular}
}
\caption{Results for valid submissions.}
\label{tab:overall_results}
\end{table}

\subsection{Top solutions}\label{sec:topsolutions}

\subsubsection{Solution 1:  }
\noindent\textit{Yi Zuo, Zitao Wang, Xiaowen Zhang, Jiaxuan Zhao, Yuting Yang, Licheng Jiao\\
yzuo\_1@stu.xidian.edu.cn
}

We used Swin-L\cite{liu2021swin}+Mask2Former\cite{cheng2022masked}, InternImage H\cite{wang2023internimage}+Mask2Former, and Segformer+MIT-B5\cite{xie2021segformer} as base models to complete the corresponding tasks.

Due to the significant category overlap between the MUAD and Cityscapes datasets, we introduced Cityscapes as a pre-trained dataset. Firstly, we trained 200000 iters on three networks on Cityscapes data to obtain corresponding pre-training weights. Subsequently, we will use the pre-trained weights obtained as initial weights and train them on the MUAD dataset. During this process, due to the lack of severe weather conditions in the MUAD test set, we designed a data augmentation method for harsh environments (AE) to simulate this situation.

In AE, we have designed new data augmentation methods for several scenarios, including rain, snow, and darkness. For example, for train data augmentation methods, we first create a random noise image to simulate raindrops. Then rotate and blur the noisy image to simulate the effect of raindrops falling. Simulate the angle and length of raindrops by rotating a diagonal matrix while applying Gaussian blur to the rotated matrix to simulate the blurred edges of raindrops. Finally, combine the raindrop effect with the original image to generate the final image with the raindrop effect. For data augmentation methods of night, we adjust the brightness, contrast, saturation, and hue of the image to simulate the effect of night. In addition, some conventional data augmentations have also been added to AE, such as Random Flip, Random Rotate, Random Crop, Random CutOut\cite{devries2017improved}, etc.

In the experiment, we found that as training increases, although mIoU gradually increases, mAUROC shows a downward trend. We believe this situation is because the model, although fitted on known categories, did not adapt to OOD categories. In order to achieve good results in mIoU and mAUROC, we use two strategies (categories that have not appeared in the train and validation sets) to address them separately. For mIoU, it relies on high accuracy in predicting category results, corresponding to "pred" in the competition. In order to improve mIoU, we need to obtain a model that can predict accurately. Therefore, we will train the three base models on MUAD with 200000 items each. Afterward, the results of the three models were ensembled. The ensemble method is as follows: (1) Use three models to predict on the test set and obtain three prediction results; (2) Vote for each pixel in the three predicted results and select the category with the highest number of votes as the final category. If it is equal, use the result of Segformer+MIT-B5; (3) Gradually iterate until the resulting voting for all test sets is completed.

For mAUROC, we use another strategy. We only fine-tune the three base networks based on Cityscapes pre-training and set a higher learning rate to prevent them from being insensitive to OOD categories. During this process, we found that the Segformer network does not clearly distinguish the boundaries of instance objects. The confidence image obtained from it presents a result similar to a noise map, but its corresponding mAUROC score is very high. For mask2Former, it corresponds to each region with a clear tendency and often clearly separates the boundary lines of the instance object. In our experiment, its score was lower than Segformer. In order to integrate high-scoring results, we selected the confidence map of Segformer as the background and fused the Mask2former results on top of the Segformer results. Simply put, we first take the reciprocal of the confidence results for Swin-L+Mask2Former and InternImage H+Mask2Former, then add them together, and then take the reciprocal as the fusion result for Mask2Former. Then, select the Mask2Former fusion result with a confidence level lower than 0.6 to directly cover the Segform confidence level result at the corresponding position.

Although this method dramatically improves the score of mAUROC, in order to pursue higher scores, we have designed a region normalization strategy. Specifically, we first use mean filtering for areas with confidence levels below 0.6, then use connected domain algorithms to segment the entire image into multiple mask regions. If more than half of the pixels have a confidence level below 0.4 in a region, we define it as an OOD region. For all pixels in the region, we cover it with the minimum confidence level of the region.

\subsubsection{Solution 2:  }
\textit{Rui Peng, Xinyi Wang, Junpei Zhang, Kexin Zhang, Fang Liu
\\22171214876@stu.xidian.edu.cn
}

\begin{figure}[t]
    \centering
    \includegraphics[width=0.5\textwidth]{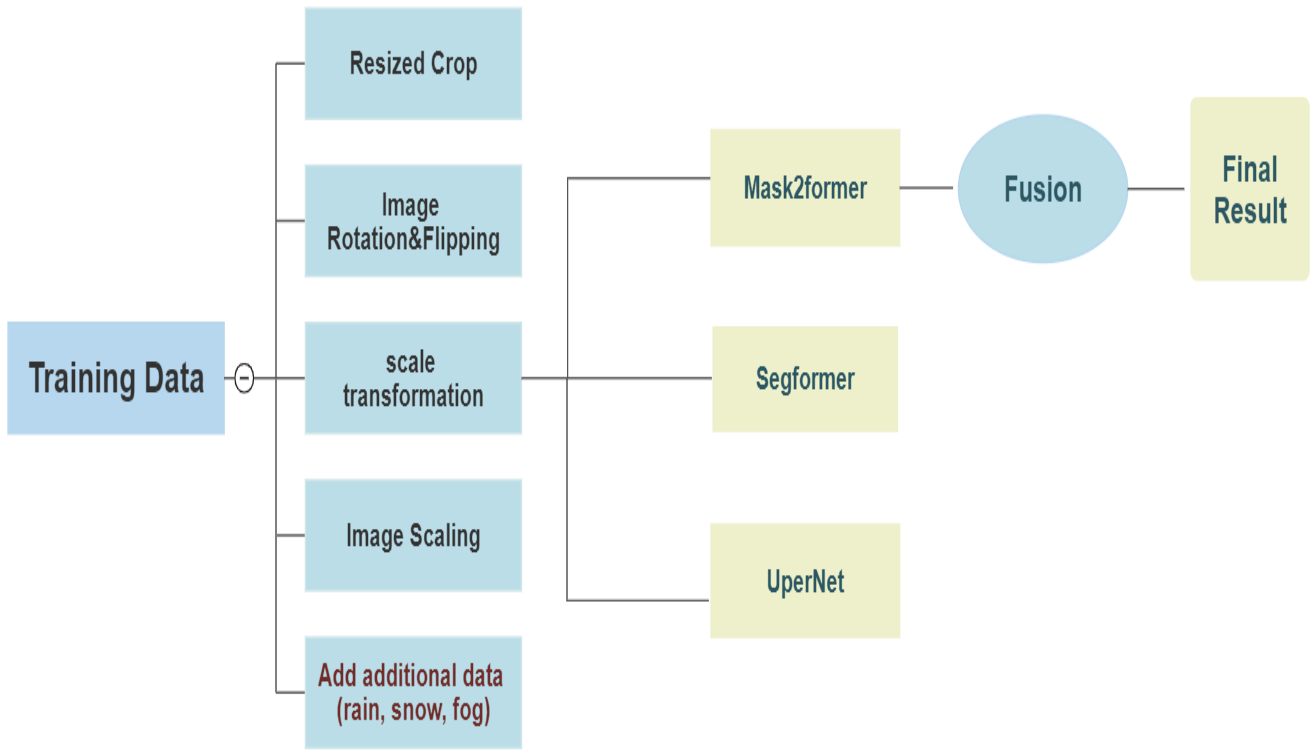}
    \caption{Overall workflow for Solution2.}
    \label{fig:solution2}
\end{figure}

The overall workflow of our proposed solution is shown in Figure~\ref{fig:solution2}. First, we train using the original dataset on three models, Mask2former~\cite{cheng2022masked}, Segformer~\cite{xie2021segformer} and UperNet~\cite{xiao2018unified}, respectively. Basic data enhancement operations, including image flipping, cropping, rotation, scaling, brightness adjustment, scale transformation, etc., were also performed on the training set. In addition, after analyzing the training, validation, and test sets, we identified data enhancement methods that can effectively improve the performance of the models. 
We added different degrees of effects of special weather conditions such as rain, snow and fog to the given training set as a way to enhance the robustness and generalization of the models. 
In the testing phase, we test the results of the three models at different stages of training to produce results. After fusing the results from multiple models, we found that the results were not optimal. 
Therefore, arithmetic averaging was finally used to fuse different results from the same model to obtain the final class prediction and confidence results to improve the segmentation effect of the model.
In the training phase, we use an image size of 512*1024 for training. For each GPU, there are 2 images per batch. In order to reduce overfitting~\cite{hoyer2023mic}, for Mask2former, we set the maximum number of iterations of training to 90000 and the initial learning rate to 6e-5 %
. For UperNet, we set the maximum number of iterations for training to 26800 with an initial learning rate of 6e-5
. For Segformer, we set the maximum number of iterations for training to 80000 and the initial learning rate to 1.2e-4
. We trained our model on a single NVIDIA RTX 3090 GPU.

\subsubsection{Solution 5: Biased Class disagreement}
\noindent\textit{Roberto Alcover-Couso, Juan C. SanMiguel, Marcos Escudero-Viñolo
\\roberto.alcover@uam.es
}

Our proposal leverages multiple models with diverse biases, aiming to assign high-confidence predictions to OOD instances by mapping them to the selected prior semantic category. These biases are introduced through the sampling strategy. Additionally, we employ soft cross-entropy loss and confidence filtering to enforce the classification of underrepresented classes. We employ an HRDA~\cite{hoyer2022hrda} architecture with the ''min pixels" parameter set to 30.  Our proposal is described in detail in the following reference~\cite{robertoiccvw2023}. 

\noindent\textbf{Teacher-student training\quad}
In order to train each model, we follow a teacher-student protocol, where two DNNs are trained: the teacher network and the student network. The teacher network is updated every time step following an exponential moving average (EMA) of the student network.

The student network is trained by minimizing a soft cross-entropy loss on the labeled images and the pseudo-labels generated by the teacher network of the unlabeled images.

\noindent\textbf{Image Sampling\quad}
Our main contribution is the employment of sampling strategies that over-sample less frequent classes based on the class frequency ($\mathbf{f}$) in the training set.
In order to oversample towards a selected class $c'$, we sample from a Bernoulli distribution, $B(1,.5)$, whether to select if a sample labeled as class $c'$ is incorporated into the training. In the case of \textit{failure}, samples are incorporated with a probability defined by the $softmax$ of the $1-\mathbf{f}$ frequencies.

\noindent\textbf{Confidence Filtering\quad}
In order to enhance the classification of low-represented classes, we propose a method that involves switching the pseudo-labels of classes exhibiting significant variation in their classification probabilities. We follow the hypothesis that pseudo-labels with a wide confidence distribution often exhibit a bimodal distribution pattern. This means that the confidence scores for false positives are generally lower and more spread out, indicating uncertainty and potential misclassifications.

Specifically, we select the 3 classes which have assigned the highest confidence standard deviation throughout the image. Then for those pixels, we filter out of the training as an unknown class the ones with the lowest confidence. We employ the mean of the confidences assigned to that class as the threshold for filtering out pixels.

\noindent\textbf{Bias class disagreement ensemble\quad}
In order to enhance the robustness of our approach, we propose to utilize an output-level ensemble  of two models. Each of the models is biased towards a different prior class through the proposed sampling. Then, we detect pixels of OOD instances by analyzing label mismatch situations under the assumption that each biased model classifies an OOD pixel as one of their respective biased class. Formally, we condition the output level ensemble by studying the pixels for which the top predicted class is the bias class of each model and consider them as pixels of OOD instances.

\subsubsection{Solution 6: }
\noindent\textit{Hanlin Tian, Kenta Matsui, Tianhao Wang, Fahmy Adan\\
h.tian22@imperial.ac.uk
}

\begin{figure}[H]
    \centering
    \includegraphics[width=0.47\textwidth]{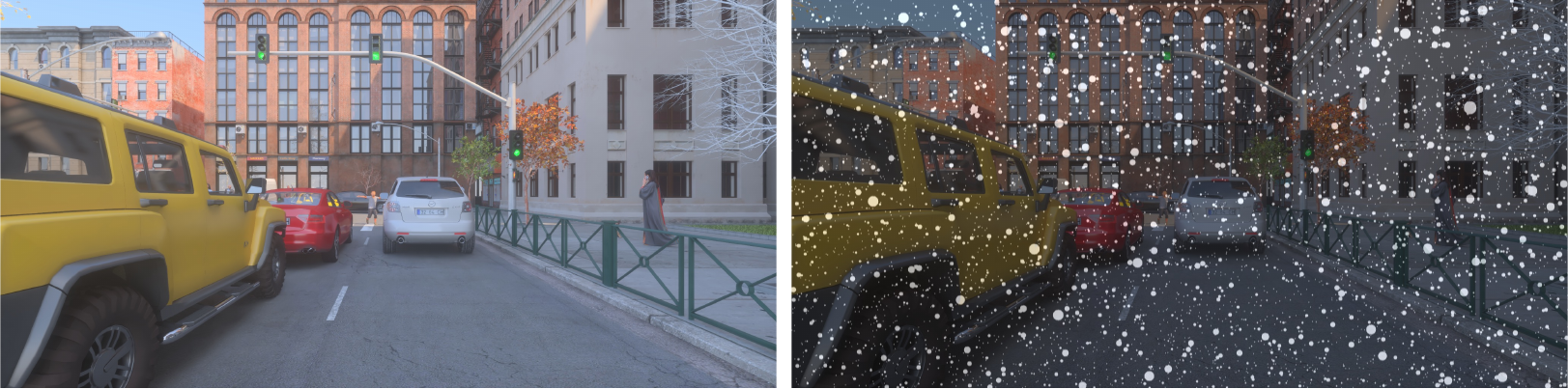}
    \vspace{0.1em}
    \caption{Data augmentation in Solution6. (Left) Original image from the training dataset. (Right) Corresponding image post-application of our weather data augmentation process.}
    \label{fig:solution6}
\end{figure}

\noindent\textbf{Data Augmentation\quad}
We introduced a data augmentation technique by employing a combination of white noise, Gaussian blur, and color alterations to simulate snowfall and its associated environmental changes. Our focus remained primarily on snowy scenarios due to time constraints. Addressing the challenges posed by rain and fog conditions remains a potential avenue for future work.

We introduced white particle-like noise to depict falling snowflakes in the data augmentation process. By varying the size of these particles, we simulated snowflakes appearing at different distances. We also introduced a slight vertical blur to mimic the motion of falling snow. We modified the color palette to colder tones in the following process, replicating the subdued and darker colors often associated with snowy scenes. We randomly selected and augmented 10\% of the entire dataset to simulate snowy conditions. Figure~\ref{fig:solution6} shows the example of the output.

\noindent\textbf{Xception Backbone\quad}
In our approach, we integrated the Xception backbone~\cite{chollet2017xception} into our segmentation model, where it serves as a feature extractor. This choice involved replacing our prior ResNet backbone~\cite{he2016deep} while retaining the DeeplabV3+ head~\cite{chen2018encoder} in our model configuration. The high-level features extracted by the Xception backbone are seamlessly fed into the subsequent segmentation layers of our model.
The incorporation of the Xception backbone led to a significant improvement in our model's overall performance. Notably, we observed that the mAUROC score, which is a pivotal metric for gauging our model’s quality, surged from 0.7861 to 0.8281. This remarkable enhancement underscores the efficacy of the Xception backbone and attests to the substantial advantages of integrating advanced architectures in confronting the challenges of intricate segmentation tasks.

\noindent\textbf{OHEM Cross-Entropy Loss\quad}
Our custom loss function, named OhemCELoss, is an implementation of Online Hard Example Mining (OHEM) with Cross-Entropy loss. This loss function is designed to focus on the hardest samples to classify, identified by a threshold on the computed loss values, and is particularly beneficial in segmentation tasks where extreme weather conditions lead to challenging pixels.

Initially, the standard Cross-Entropy loss between the logits and ground truth labels is computed without reduction:

\begin{equation}
    L_i = - \log \left(\frac{{\exp(x_{i,y_i})}}{{\sum_j \exp(x_{i,j})}}\right)
\end{equation}\noindent
where $L_i$ is the loss for the $i$-th sample, $x_{i,j}$ represents the logits, and $y_i$ is the true label for the $i$-th sample.
    
Hard examples are identified as those samples where the loss exceeds a dynamically computed threshold (thr):
\begin{equation}
    \text{thr} = \max \left( \text{thr}, \min \left( L_{(\text{sorted})}[\text{min\_kept}], L_{(\text{sorted})}[\text{end}] \right) \right)
\end{equation}\noindent
where $L_{(\text{sorted})}$ is the sorted vector of losses.
    
The final OHEM loss is the mean loss value over these selected hard examples:

\begin{equation}
    \text{OHEM Loss} = \frac{1}{N} \sum_{i=1}^{N} L_i \cdot \mathbb{I}(L_i > \text{thr})
\end{equation}\noindent
where $N$ is the total number of examples, and $\mathbb{I}(\cdot)$ is the indicator function, which is $1$ if the condition inside is true, and $0$ otherwise.

In practice, this results in the model focusing more intently on the most challenging parts of the images—often, these are regions with intricate textures, object boundaries, or extreme weather effects. By setting an ignore label parameter (e.g., to $255$), we explicitly instruct the model to disregard certain regions, which is useful for handling out-of-distribution (OOD) areas in the segmentation task.

\subsubsection{Solution 7: Energy Score with BN adaptation}
\noindent\textit{Zhitong Gao, Xuming He
\\gaozht@shanghaitech.edu.cn
}

The method adopts the energy score as the base strategy for OOD object detection, which aligns theoretically with the probability density of the inputs and offers better performance than softmax scores~\cite{liu2020energy}. However, applying energy scores directly as a post-hoc approach often suffers from the presence of special weather conditions in the test set.

To overcome this challenge, the method introduces a novel instance-wise test-time domain adaptation technique, which enables the backbone network adapts the batch normalization (BN) layers according to image statistics at test time. 
Specifically, the overall framework operates in two sequential phases: 
first, it detects and mitigates the effects of unknown weather conditions through the adaptive BN; next, it performs OOD detection by computing a pixel-wise energy score map of the model's outputs. The entire process unfolds through the following specific steps:

\noindent\textbf{Model Training\quad}
The method does not modify the training process and can use any standard closed-world segmentation training procedures. In this work, a pretrained Deeplab v3+ with Resnet 101 model, provided by the official benchmark GitHub repository~\footnote{\url{https://github.com/ENSTA-U2IS/DeepLabV3Plus-MUAD-Pytorch}}, is used.%

\noindent\textbf{BN Adaptation\quad}
When exposed to online test data, the model employs a test-time adaptation module to identify and adapt to special weather conditions that might be present in the data. Inspired by the technique of transductive batch normalization~\cite{TBN}, we implement the adaptation by updating the BN statistics of the network. Specifically, the running mean and standard deviation $\mu^{r}, \sigma^{r}$, are mixed with the mean and standard deviation $\mu{(x)}, \sigma(x)$ for each test input $x$.  The updated mean and standard deviation are computed as $\hat{\mu} = \alpha * \mu(x) + (1-\alpha) * \mu^{r}$, $\hat{\sigma}_l^2 = \alpha * \sigma^2(x) + (1-\alpha)* (\sigma^{r})^2$. Here, $\alpha$ is the mixing coefficient, representing the probability that the input image originates from an unknown weather condition.  This coefficient is estimated by considering the Kullback–Leibler (KL) divergence between the feature distributions of the test image and the training images within the Normalization layers: $KL(N(\mu(x), \sigma(x)) || N(\mu^{r}, \sigma^{r}))$. The mixing coefficient $\alpha$ is calculated by averaging the KL divergence values over all BN layers and normalizing the result to a 0-1 range using the sigmoid function.

\noindent\textbf{Prediction and OOD detection\quad}
After the BN Adaptation, the method proceeds to compute the segmentation network's predictions along with an energy score as the basis for OOD detection. Specifically, the logits of the model output are denoted as $\eta_i$, where $i$ identifies the pixel.  The inlier prediction is obtained through: $\argmax_{c}{\exp({\eta_i^c})} / {\sum_{n}^{N_c}} \exp({\eta_i^n})$, and the energy score is calculated using the energy function: $\log\sum_{n}^{N_c} \exp({\eta_i^n})$. This score represents a quantitative measure of the model's certainty regarding its predictions and is employed to detect out-of-distribution (OOD) instances within the test data. 
    
Overall, the method synergizes the energy scores with a BN adaptation strategy to address the unique challenges posed by special weather conditions, which provides a robust and effective solution for OOD detection in the wild.

\subsubsection{Solution 8: Polynomial Calibration}
\noindent\textit{Quentin Bouniot, Hossein Moghaddam
\\quentin.bouniot@telecom-paris.fr
}

We apply our technique on the baseline model, i.e. DeepLabV3+~\cite{chen2018encoder} with ResNet101 backbone~\cite{he2016deep}. Among the techniques employed to quantify uncertainty, one that consistently demonstrates commendable performance involves the estimation of the posterior distribution $P(\vtheta|\mathcal{D})$ of the network \cite{blundell2015weight,maddox2019simple,franchi2020tradi}. Subsequently, $N$ DNNs are sampled according to this posterior, denoted as ${ F_{\vtheta_j}(\cdot) }_{j=1}^N$, and their performance is averaged as follows:

\begin{equation}
    \hat{\vy}_i = P(Y=\vy_i | X=\vx_i ) = 1/N \sum_j F_{\vtheta_j}(\vx_i)
\end{equation}

The Deep Ensembles \cite{lakshminarayanan2017simple} (DEns) technique simplistically estimates this posterior, often yielding impressive results due to its capability, highlighted in \cite{wilson2020bayesian}, to estimate diverse modes of the posterior. However, the training of multiple DNNs makes DEns challenging. Various strategies have been proposed to streamline DEns training \cite{wen2020batchensemble, durasov2021masksembles,havasi2020training,laurent2023packed}, but due to the time-intensive nature of training, the number of samples for ensemble training remains limited. Alternatively, the Monte Carlo (MC) dropout \cite{gal2016dropout} method estimates a singular mode of the posterior, offering a more accessible solution as it requires only one training. Consequently, this approach can furnish a larger sample set. However, both these techniques introduce increased complexity to the inference process.

Smith and Gal \cite{smith2018understanding} interestingly amalgamated these two Bayesian Neural Network approximations to develop Dropout Ensembles (DropEns), reaping the benefits of both preceding techniques. Nonetheless, DropEns may encounter calibration prediction issues. Several calibration techniques for trained DNN predictions have been explored~\cite{shih2023long,rahimi2020post,khalifa2021improving} to address this issue, commonly referred to as post-hoc calibration. Among these, we evaluated temperature scaling~\cite{guo2017calibration} (TS), a method involving a temperature parameter $\tau \geq 0$ to the DNN logits. Specifically, if we denote $F_{\vtheta_j}(\vx_i) = \mbox{soft} \left( G_{\vtheta_j}(\vx_i) \right)$, where $\mbox{soft} (\cdot)$ represents the softmax function and $G_{\vtheta_j}(\cdot)$ signifies the logit of $F_{\vtheta_j}(\cdot)$, temperature scaling entails tuning $\tau$ to achieve the optimal ECE for $F_{\vtheta_j}^{\tau}(\vx_i) = \mbox{soft} \left( G_{\vtheta_j}(\vx_i)/ \tau \right)$.

Typically, any monotonically increasing function can be employed for this purpose. In our approach, we use a polynomial function, considering that we normalize the logits by subtracting the minimum value for each data point to ensure their positivity. Thus, the normalized logit becomes $G^{\mbox{norm}}_{\vtheta_j}(\cdot)$. We then apply the polynomial temperature scaling (PTS) function incorporating parameters $\tau_1$, $\tau_2$, and $\tau_3$, leading to the following DNNs:

\begin{multline}
    F_{\vtheta_j}^{\tau_1,\tau_2,\tau_3}(\vx_i) = \mbox{soft}  \left( G^{\mbox{norm}}_{\vtheta_j}(\vx_i)/ \tau_1  \right. \\
    \left. + (G^{\mbox{norm}}_{\vtheta_j}(\vx_i)/ \tau_2)^2  + (G^{\mbox{norm}}_{\vtheta_j}(\vx_i)/ \tau_3)^3  \right)
\end{multline}

An overview of our findings is presented in Tab~\ref{tab:solution8}. It's evident that DEns and standalone MC Dropout exhibit commendable performances. However, they fall short of effectively capturing the underlying uncertainty inherent in the DNNs. Notably, the DropEns outperforms its counterparts in terms of performance, and its calibrated version further enhances its effectiveness.

\begin{table}[t!]
\centering
\scalebox{0.67}{
\begin{tabular}{lccccc} 
\toprule
Technique & mAUROC~$\uparrow$ & mAUPR~$\uparrow$ & mFPR~$\downarrow$ & mECE~$\downarrow$ & mIoU$\uparrow$ \\ 
\toprule
Baseline & 0.7337 & 0.1790 & 0.5253 & 0.2880 & 0.3690 \\ 
\midrule
MC Dropout & 0.7794 & 0.1778 & 0.4791 & 0.1380 & 0.4358 \\ 
\midrule
DEns & 0.7399 & 0.1884 & 0.4787 & 0.2521 & 0.3764 \\ 
\midrule
DropEns & 0.8195 & 0.2243 & 0.4015 & 0.1093 & \second 0.4547 \\ 
\midrule
DropEns+TS & \second 0.8219 & \second 0.2261 & \second 0.3988 & \second 0.1005 & 0.4544 \\ 
\midrule
DropEns+PTS & \first \textbf{0.8356} & \first 
 \textbf{0.2502} & \first \textbf{0.3898} & \first \textbf{0.0942} & \first \textbf{0.4558} \\
\bottomrule
\\
\end{tabular}
}
\caption{Ablation study for Solution8.}
\label{tab:solution8}
\end{table}

\subsubsection{Solution 9: Uncertainty Calibrated Mask2Former}
\noindent\textit{Shyam Nandan Rai, Fabio cermelli and Carlo Masone
\\shyam.rai@polito.it 
}

Mask architectures~\cite{cheng2021maskformer,cheng2022masked} have proven to be an effective method for universal segmentation and anomaly segmentation~\cite{shyam2023M2A,nayal2023RbA}. However, these networks are not well calibrated, resulting in poor uncertainty estimates. Hence, we calibrate the Mask2Former architecture~\cite{cheng2022masked} to give better uncertainty estimates and identify OOD objects. For this purpose, we introduce the following modifications to the architecture: a) during training, we change the loss function from cross-entropy to focal loss~\cite{lin2017focal} as it allows us to learn well-calibrated models, and b) at inference, we propose a way to perform temperature scaling in mask-architecture. After integrating modification into mask architecture, we call the solution Uncertainty Calibrated Mask2Former (UC-M2F). It is also important to note that our method uses no additional OOD training data. In the subsequent sub-sections, we will delve into each of these modifications.

\noindent\textbf{Preliminary\quad}
Mask2Former consists of three major parts: a) a \textit{backbone} acting as a feature extractor, b) a \textit{pixel-decoder} that upsamples the low-resolution features from the backbone to produce high-resolution \textit{per-pixel embeddings}, and c) a \textit{transformer decoder}, that takes the image features to output a fixed number of object queries consisting of \textit{mask embeddings} and their associated \textit{class scores} $C\in \mathbb{R}^{N \times N_{C}}$. The final \textit{class masks} $M\in \mathbb{R}^{N\times (H \times W)}$ are obtained by multiplying the mask embeddings with the per-pixel embeddings obtained from the pixel-decoder. $N$ represents the number of output masks.

\noindent\textbf{Training Loss\quad}
Mask2Former is trained with a cross-entropy loss to predict $C$ and both a binary cross-entropy and 
a dice loss to predict $M$. 
However,~\cite{mukhoti2020calibrating} points out that training deep networks on standard cross-entropy loss results in miscalibrated network and suggests using focal loss~\cite{lin2017focal} can lead to a very well-calibrated network. So, we change the training loss for the class scores to focal loss

\begin{equation}
    L_{fl}(x_{i}) = - \alpha (1 - \hat{y}^{i})^\gamma \log(\hat{y}^{i})
\end{equation}\noindent
where $\alpha$ and $\gamma$ are the scaling and focusing factors, respectively.

\noindent\textbf{Temperature scaling\quad}
Temperature scaling 
is another way to improve the calibration of a network without affecting accuracy. However, temperature calibration in Mask2Former is not trivial due to the nature of its output calculation given:
\begin{equation}
\label{eq:output}
g(x_{i}) = \max^{N_{C}} \left(\text{softmax}(C)^T \cdot \text{sigmoid}(M)\right)
\end{equation}\noindent
We now have three options to calibrate the network: dividing $C$ by temperature, dividing $M$ by temperature, or both. We found that dividing $C$ by the temperature gives the best results. Formally, eq.~\ref{eq:output} becomes:
\begin{equation}
g(x_{i}) = \max^{N_{C}} \left(\text{softmax}(C/t)^T \cdot \text{sigmoid}(M)\right)
\end{equation}\noindent

\noindent\textbf{Training Details\quad}
Our UC-M2F architecture uses Swin-B as the feature backbone, and the decoder is kept the same as Mask2Former. The network is trained
with an initial learning rate of 1e-4 and batch size of 8 for
90 thousand iterations on AdamW with a weight decay
of 0.05. Then, fine-tuned for 5 thousand iterations having a learning rate of 1e-5. We use an image crop of 400 $\times$ 800 with large-scale jittering along with a random scale ranging from 0.1 to 2.0. In the focal loss, $\gamma$ is kept at 2, and $\alpha$ is 10. The best temperature value is found to be 0.5.